\documentclass[10pt,titlepage]{report}
\usepackage{Preamble}
\usepackage{multirow}

\hypersetup{colorlinks = true, urlcolor = blue, linkcolor = red, citecolor = green}

\begin{document}

\begin{titlepage}
    \newgeometry{margin=3cm}
	\centering
    \includegraphics[width=0.4\linewidth]{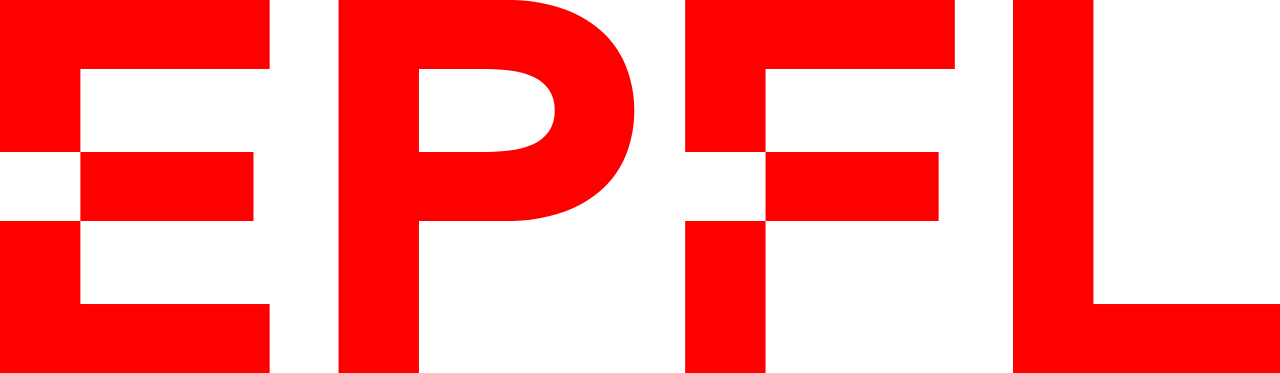}
    \\[0.25cm]
    \textsc{\large École Polytechnique Fédérale de Lausanne}\\ \vspace{\fill}
    \textbf{\textsc{\fontsize{40}{40}\selectfont Report}}\\\vspace{1.5in}
    \textsc{\LARGE Learning degraded image classification with restoration data fidelity}\\
    \vspace{\fill}		
	\textsc{\LARGE Semester project \vspace{10pt}\\Image and Visual Representation Lab (IVRL)}\\[0.4cm]
	\rule{\linewidth}{0.2 mm} \\[0.5 cm]
	Student: Xiaoyu Lin\\Supervisors: Majed El Helou, Deblina Bhattacharjee\\Professor: Sabine S\"usstrunk\\[2cm] \today
\end{titlepage}
\restoregeometry

\thispagestyle{numberonly}
\pagenumbering{roman}


\chapter*{Abstract}
    \addcontentsline{toc}{chapter}{Abstract}
    Learning-based methods especially with convolutional neural networks (CNN) are continuously showing superior performance in computer vision applications, ranging from image classification to restoration. 
For image classification, most existing works focus on very clean images such as images in Caltech-256 and ImageNet datasets. 
However, in most realistic scenarios, the acquired images may suffer from degradation. 
One important and interesting problem is to combine image classification and restoration tasks to improve the performance of CNN-based classification networks on degraded images. In this report, we explore the influence of degradation types and levels on four widely-used classification networks, and the use of a restoration network to eliminate the degradation's influence. We also propose a novel method leveraging a fidelity map to calibrate the image features obtained by pre-trained classification networks. We empirically demonstrate that our proposed method consistently outperforms the pre-trained networks under all degradation levels and types with additive white Gaussian noise (AWGN), and it even outperforms the re-trained networks for degraded images under low degradation levels. We also show that the proposed method is a model-agnostic approach that benefits different classification networks. Our results reveal that the proposed method is a promising solution to mitigate the effect caused by image degradation.


\tableofcontents


\chapter{Introduction}
    \pagenumbering{arabic}
    \section{Background}

Neural networks have been shown to substantially outperform traditional approaches in various fields, ranging from pattern recognition to robotics. 
The popularity of deep learning in the field of image recognition has grown exponentially in the past decade after deep convolutional neural networks (CNN) boosted its performance~\cite{krizhevsky2012imagenet} on this task. 
Nowadays, CNN-based methods have grown from a specialist niche to become mainstream for virtually every computer vision and image processing task known to the community. 
Much research has focused to define a new architecture that gets better performance on well-known image classification challenge and datasets~\cite{szegedy2015going, ioffe2015batch, Szegedy_2016_CVPR, simonyan2014very, he2016deep}. 
A large effort was also made to modify those popular architectures to be better suited to other tasks such as image restoration~\cite{zhang2017beyond,el2020blind}.

Although the CNN-based methods have boosted their performance on both high-level image classification problems~\cite{krizhevsky2012imagenet} and low-level image restoration problems~\cite{zhang2017beyond}, combining them, for example, exploring the performance of CNN-based classification network on degraded images, has attracted less attention. However, this problem is also very interesting and important as acquired images are always degraded in most practical scenarios, due to sensor noise~\cite{winkler2004visibility}, variable focal blur~\cite{el2018aam}, light conditions~\cite{dumbgen2018near}, human factors such as motion~\cite{ji2008motion}, etc. And because restoration networks are capable of various hallucinations both spatially~\cite{elhelou2020bigprior} and in terms of frequency components~\cite{elhelou2020stochastic}, making their direct application ahead of the classification network~\cite{liu2019classification} far less beneficial than can be expected.

The methods that have been proposed to help classification networks overcome the problem caused by degradation can be divided into two main types. The first one solve this problem by using transfer learning. Therefore, they employ popular methods in transfer learning such as fine-tuning~\cite{peng2016fine}, layer sharing \cite{wang2016studying, ghosh2018robustness}, multi-branch ensemble \cite{endo2020classifying} and domain adaptation \cite{sindagi2020prior}. Those methods retrain the whole or part of the original classification network, which is very complex and computationally-expensive. Another alternative is much simpler, namely to use restoration networks as a pre-processing step and pass the restored images to classification network without changing the classification network. However, these methods assume that the restoration and classification network have a consistent optimization target, which is not true in most cases, leading to suboptimal results.

We propose to address the problem, \textit{without} going through the computationally-expensive classification network re-training. To that end, we make use of a fidelity map to calibrate the image feature extracted by classification networks in the presence of image degradation. The fidelity map we use is the pixel-wise $\ell_1$ distance between the restored image and clean image. In this report, we propose a novel method to use the fidelity map, which works like an attention mechanism to make the network pay more attention to less degraded pixels, and apply it on both channel and spatial dimensions. The proposed method can calibrate the image features obtained by pre-trained classification networks without changing the architecture or parameters of the pre-trained network. In this case, we can make full use of pre-trained networks and save on computational cost and model complexity while mitigating the drop in accuracy that is normally caused by the degradation problem. 

\section{Objectives}

The aim of the project is to develop a novel approach for classification networks pre-trained on clean images to mitigate the effect caused by image degradation. More specifically, the main target of this project is to design a model-agnostic approach for CNN-based classification networks under some specific degradation types such as additive white Gaussian noise (AWGN).

To achieve this target, the following objectives are identified:
\begin{itemize}
\setlength\itemsep{-0.5em}
    \item To conduct a literature survey on the current state-of-the-art degraded image classification networks, including image restoration networks.
    \item To explore the effects of degradation types and levels on classification networks.
    \item To explore the effects of fidelity map on pre-trained classification networks when suffer from image degradation. 
    \item To show the generalization ability of the proposed method.
\end{itemize}

\section{Report organization}

This report consists of eight chapters in the main body, and it is structured as follows.
	
After this introductory chapter, the second chapter briefly reviews the history of CNN-based classification networks and some methods to help those networks to perform better on degraded images.

Some experiments are designed to explore the effects of degradation types and levels on classification networks. The results of these tests are introduced in the third chapter and guide our research in the following chapters.

We propose a method based on fidelity map which is the pixel-wise $\ell_1$ distance between restored images and clean images in the fourth chapter. 

In the fifth chapter, we empirically demonstrate that the proposed method can greatly improve the performance of pre-trained classification networks on degraded images. 

In the sixth and seventh chapter, we show that all modules in the proposed method contribute to the improvements, and the proposed method is a model-agnostic approach, respectively.

In the last chapter, the conclusion of the proposed method, as well as some plans and suggestions for further research, are introduced.

\chapter{Literature review}
    The last few years have witnessed more and more deep learning techniques applied in the field of computer vision. As one of the most classical topics in computer vision, image classification benefits a lot from those methods especially CNN~\cite{lecun1998gradient}. This chapter firstly gives a brief overview of the development of CNN-based image classification networks on clean natural images. Then some attempts to adapt those classification networks on degraded images are introduced. Since some of the methods~\cite{endo2020classifying, sharma2018classification} apply image restoration to degraded images as a pre-processing step before classification, we also review some state-of-the-art image restoration methods in this chapter.

\section{Clean natural image classification networks}
Since 2012 deep CNN started to become mainstream at the core of most state-of-the-art computer vision solutions for image classification tasks.
AlexNet~\cite{krizhevsky2012imagenet} is the first CNN-based model applied in the large-scale high-resolution image recognition task and dramatically outperforms traditional methods. 
Inspired by 1x1 convolution in NIN~\cite{lin2013network}, GoogLeNet~\cite{szegedy2015going} proposes an Inception module to capture information at different scales in the same stage with controlled computation complexity, which is further improved by batch normalization~\cite{ioffe2015batch} and label smoothing~\cite{Szegedy_2016_CVPR}. 
VGG~\cite{simonyan2014very} increases the depth of their network with very small convolution filters. 
Although the performance of CNN-based network can be improved by increasing its depth and width, the performance will reach saturation~\cite{simonyan2014very} due to limited dataset size and model capability. 
ResNet~\cite{he2016deep} proposes a residual block by using a skipping connection to build a much deeper network while eliminates gradient vanishing problems. 
After that, more CNN-based networks are proposed and get significant progress, e.g. DenseNet~\cite{huang2017densely}, MobileNet~\cite{howard2017mobilenets} and its variations~\cite{sandler2018mobilenetv2, howard2019searching}, etc. 

\section{Degraded image classification}
Although the aforementioned methods achieve impressive results, most of them are designed and tested on clean natural image datasets such as ImageNet~\cite{deng2009imagenet}. 
However, this is not available in most practical scenarios, where images are always degraded due to sensor noise, light conditions, and human factors, etc. 
Recent studies~\cite{dodge2016understanding, pei2019effects, roy2018effects} have shown that degradation significantly reduces the accuracy of image classification especially when the gradation levels and types are largely mismatched between training and testing images. 
This is further proved by~\cite{tadros2019assessing} finding that CNN-based classification networks are more susceptible to degradation than human observers. 
Therefore, some methods have been proposed to help classification networks overcome the problem caused by degradation. 
Those methods fall into two major categories.

On the one hand, the clean and degraded images can be regarded as data from different domains, therefore the task to classify degraded images can be viewed from a transfer learning perspective. 
\cite{peng2016fine} continually fine-tunes the model trained on high-resolution images while gradually changes training data to the low-resolution domain. 
Layer sharing is also commonly used in transfer learning tasks. 
\cite{wang2016studying} partially shares CNN filters between low and high-resolution channels. \cite{ghosh2018robustness} trains three slave CNN branches according to degradation levels and uses a master CNN to select a specific branch for classification. Using a similar idea of a multi-branch network, \cite{endo2020classifying} proposes an ensemble network consisting of a restoration network followed by two classification networks: one for clean images and the other for degraded images. An estimation network is designed to estimate degradation levels and then ensemble weights. 
\cite{sindagi2020prior} views the removal of the degradation caused by weather conditions as an unsupervised domain adaptation task. 
Since the degradation can be mathematically modeled, \cite{sindagi2020prior} uses this prior knowledge to formulate a new prior-adversarial loss to capture weather-invariant features.

On the other hand, image restoration or enhancement methods can be applied to degraded images as a pre-processing step to improve classification performance. \cite{sharma2018classification} proposes a unified dynamic CNN architecture that emulates a range of enhancement methods to improve both classification and enhancement performance. 

We also review some image restoration methods here.
Because CNN-based method cannot extract global features when the input image is large or the network itself is too simple, \cite{kinoshita2019convolutional} adds a global encoder and concatenates the extracted feature with the original local one. 
Inspired by the adjustable receptive field in human vision systems, MIRNet~\cite{zamir2020learning} introduces a new feature extractor that captures both spatial and contextual details by exchanging knowledge from multi-scale features. 
The performance of the image restoration task is substantially improved by Generative Adversarial Network (GAN). To capture richer prior from large-scale and more complex natural images by GAN, DGP \cite{pan2020exploiting} relaxes the constrain of the traditional GAN-inversion problem by allowing the generator to be fine-tuned along with the latent vector. 
The discriminator feature distance metric is introduced to keep generative prior, and a progressive reconstruction procedure is proposed to reduce the mismatch between low-level and high-level information.
ForkGAN~\cite{zheng_2020_ECCV} proposes a fork-shaped generator containing one encoder and two decoders (reconstruction decoder and translation decoder) to disentangle the domain-specific and domain-invariant information.

Although the aforementioned methods all aim to improve the classification accuracy of CNN-based classification networks, they implement different classification networks in their original papers.
We summarize classification networks used in related work above and list them in Table~\ref{tab:classification_networks}. 
\begin{table}[ht]
    \small
    \centering
    \begin{tabular}{|c|c|}
        \hline
        Related work
        & Classification network(s)
        \\
        
        \hline
        \cite{pei2019effects}
        & AlexNet~\cite{krizhevsky2012imagenet}, 
        VGG-16~\cite{simonyan2014very}, 
        ResNet-50~\cite{he2016deep}\\
        
        \hline
        \cite{roy2018effects}
        & MobileNet~\cite{howard2017mobilenets}, 
        VGG-16~\cite{simonyan2014very}, 
        VGG-19~\cite{simonyan2014very}, 
        ResNet-50~\cite{he2016deep}, 
        Inception-V3~\cite{Szegedy_2016_CVPR}, 
        CapsuleNet~\cite{sabour2017dynamic}\\
        
        \hline
        \cite{tadros2019assessing}
        & AlexNet~\cite{krizhevsky2012imagenet}\\
        
        \hline
        \cite{endo2020classifying}
        & VGG-like~\cite{simonyan2014very}\\

        \hline
        \cite{sharma2018classification} 
        & AlexNet~\cite{krizhevsky2012imagenet}, 
        GoogLeNet~\cite{szegedy2015going}, 
        VGG-VD~\cite{simonyan2014very}, 
        VGG-16 \cite{simonyan2014very}, 
        BN-Inception \cite{ioffe2015batch}\\ 
        
        \hline
        \cite{peng2016fine}
        & AlexNet~\cite{krizhevsky2012imagenet}\\
        
        \hline
        \cite{wang2016studying}
        & AlexNet-like~\cite{krizhevsky2012imagenet}\\
        
        \hline
        \cite{ghosh2018robustness}
        & VGG-16~\cite{simonyan2014very}, 
        AlexNet~\cite{krizhevsky2012imagenet}\\
        
        \hline
        \cite{sindagi2020prior}
        & VGG-16~\cite{simonyan2014very} (Faster-RCNN, Object detection)\\
        
        \hline
    \end{tabular}%
    \caption{Summary of classification networks used in related work}
    \label{tab:classification_networks}%
\end{table}%
    
\chapter{Problem definition \& analysis}
    \label{cha:problem}
    In this chapter, we explore the effects of degradation types and levels on classification networks. We first introduce classification networks used in this report in Section~\ref{sec:classification}. Then we propose three experiment setups to train and evaluate those classification networks separately in Section~\ref{sec:effect}. Finally, we summarise and analyses our results in Section~\ref{sec:analysis} for further research.

\section{Classification networks}
\label{sec:classification}

For a comprehensive analysis of the effects of image degradation on CNN-based classification networks, we implement AlexNet~\cite{krizhevsky2012imagenet}, GoogLeNet~\cite{szegedy2015going}, VGG-16~\cite{simonyan2014very}, ResNet-50~\cite{he2016deep} in \textit{torchvision}\footnote{ https://pytorch.org/docs/stable/torchvision/models.html}, and test their performance under different types and levels of degradation. 
All classification networks are pre-trained on the ImageNet dataset. 
We use Caltech-256\footnote{http://www.vision.caltech.edu/Image\_Datasets/Caltech256} for training and evaluation in this report. 
The Caltech-256 dataset contains 256 object categories and 1 clutter category with 30607 images in total. We follow \cite{pei2019effects} to use all 257 categories. 
For each class, we randomly select 60 images as the training set, and randomly select 20\% from training set as the validation set. 
The rest images are used for testing.

During training, we perform the following steps one-by-one~\cite{he2019bag}:
    \begin{enumerate}
        \setlength\itemsep{-0.5em}
        \item Randomly sample an image and decode it into 32-bit floating point raw pixel values in $[0.0, 1.0]$.
        \item Randomly crop a rectangular region whose aspect ratio is randomly sampled in $[3/4, 4/3]$ and are randomly sampled in $[8\%, 100\%]$, then resize the cropped region into a 244-by-244 square image.
        \item Flip horizontally with 0.5 probability.
        \item Normalize RGB channels by subtracting $[0.485, 0.456, 0.406]$ and dividing by $[0.229, 0.224, 0.225]$.
    \end{enumerate}
During validation and testing, we resize each image's shorter edge to 224 pixels while keeping its aspect ratio. Next we crop out the 224-by-224 region in the center and normalize RGB channels similar to training.

The final fully-connected layers in all classification networks are modified to fit the number of classes of Caltech-256 dataset (i.e. 257). 
We also modify the final fully-connected layers in the GoogLeNet auxiliary loss.  
Then all classification networks are fine-tuned on our training set before testing. 
For fine-tuning, we adapt the training procedure in \cite{he2019bag} and apply the same procedure to all four networks. 
The weights of modified fully-connected layers are initialized with the Xavier algorithm \cite{glorot2010understanding}, and the biases are initialized to 0. 
We use Nesterov Accelerate Gradient (NAG) descent \cite{nesterov1983method} optimizer with an initial learning rate of 0.001 and a batch size of 64. 
The total number of training epochs is set to be 120. 
We also introduce a batch-step linearly learning rate warmup \cite{goyal2017accurate} for the first 5 epochs and cosine learning rate decay \cite{he2019bag}. 
We apply label smoothing \cite{Szegedy_2016_CVPR} with $\varepsilon=0.1$, as this trick is shown in \cite{he2019bag} to improve the performance of classification. 
For each epoch, we calculate and report the loss and accuracy on training and validation sets (Fig.~\ref{fig:finetune-clean}). 
Finally, we choose the model with the highest accuracy on the validation set among all training epochs for further testing and analysis.

\section{Effects of degradation on classification networks}
\label{sec:effect}

\begin{figure}
    \centering
    \includegraphics[width=.7\linewidth]{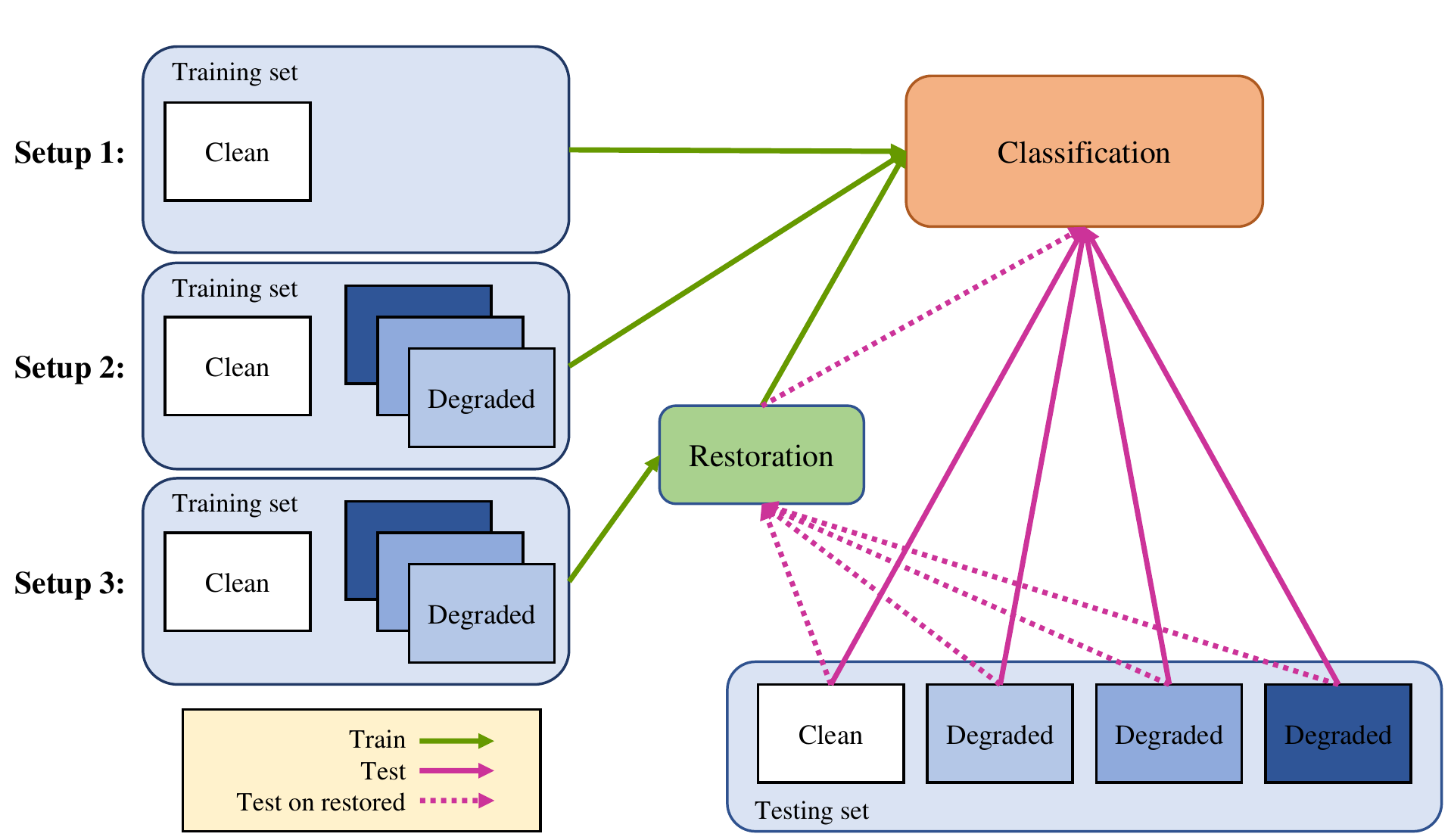}
    \caption{Experiments setup}
    \label{fig:setup}
\end{figure}

In order to find the effects of degradation types and levels on classification networks, we designed a series of experiments.
Those experiments are mainly divided into three setups as illustrated in Fig.~\ref{fig:setup}. 
The following sections will introduce those three setups and corresponding results.

\subsection{Setup 1: train on clean images}
\label{sub:clean}
\begin{figure}
    \centering
    \includegraphics[width=\linewidth]{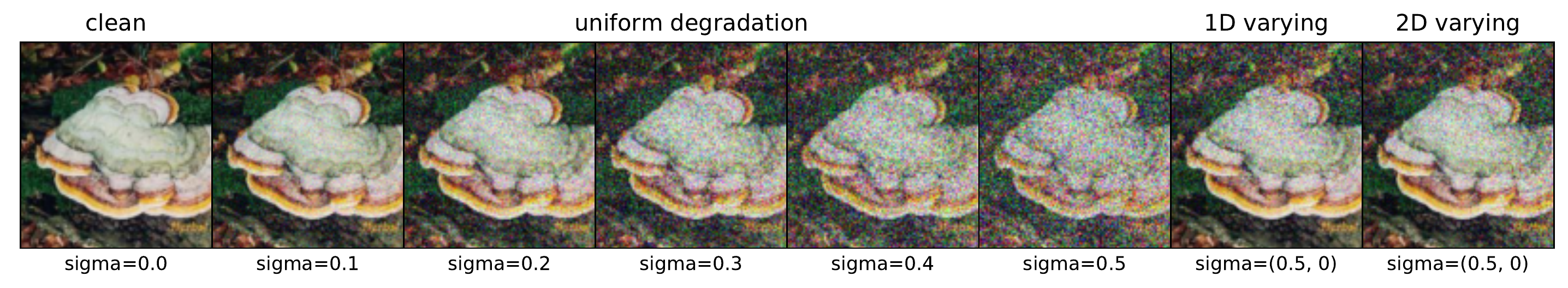}
    \caption{Examples of the synthesized noisy images with AWGN. For spatially varying degradation, $sigma=(0.5, 0)$ means the degradation level varies linearly from $0$ to $0.5$.}
    \label{fig:degradation}
\end{figure}

\begin{figure}
    \centering
    \begin{subfigure}[b]{0.32\textwidth}
        \centering
        \includegraphics[width=\textwidth]{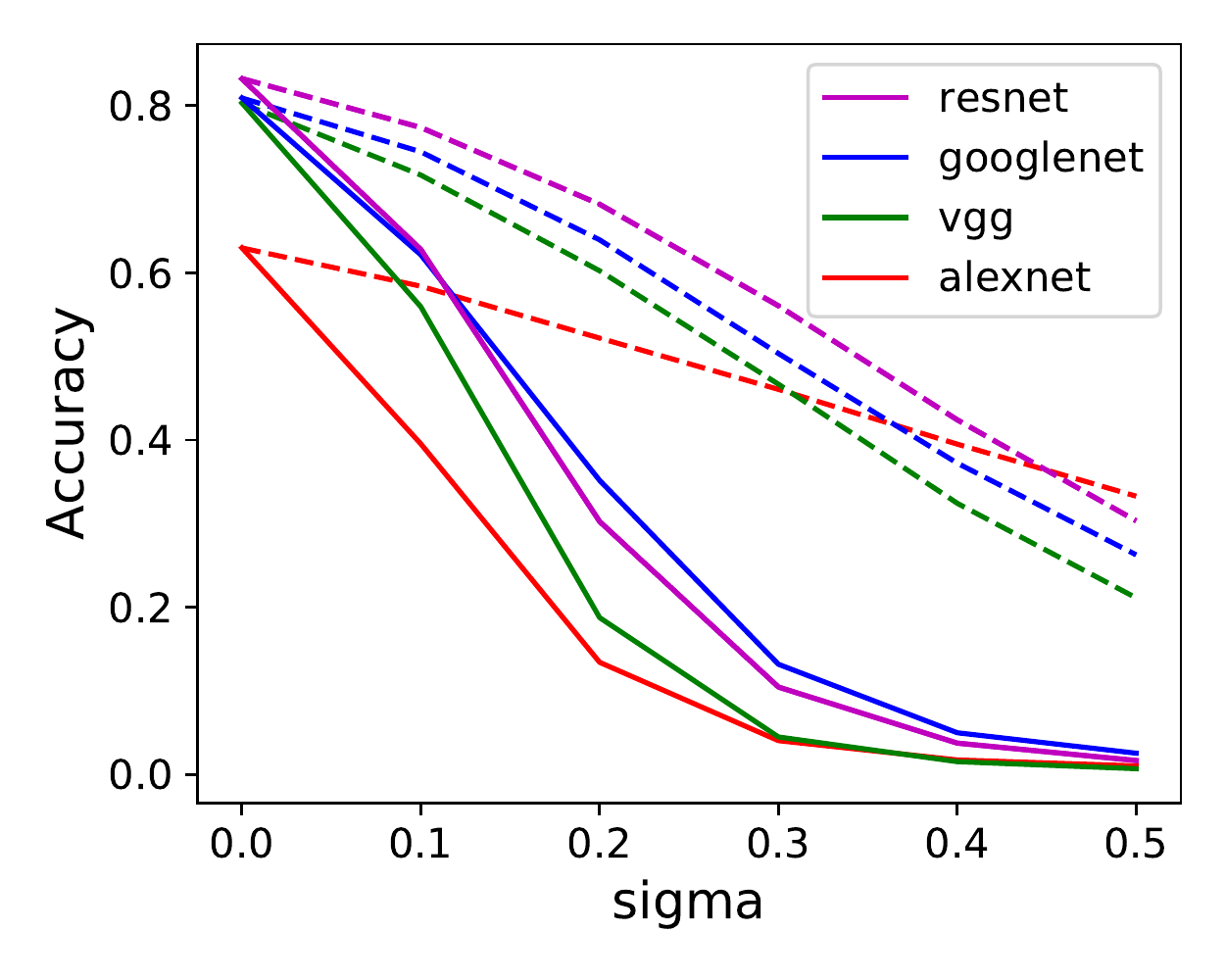}
        \caption{Setup 1 results}
        \label{fig:setup1}
    \end{subfigure}
    \hfill
    \begin{subfigure}[b]{0.32\textwidth}
        \centering
        \includegraphics[width=\textwidth]{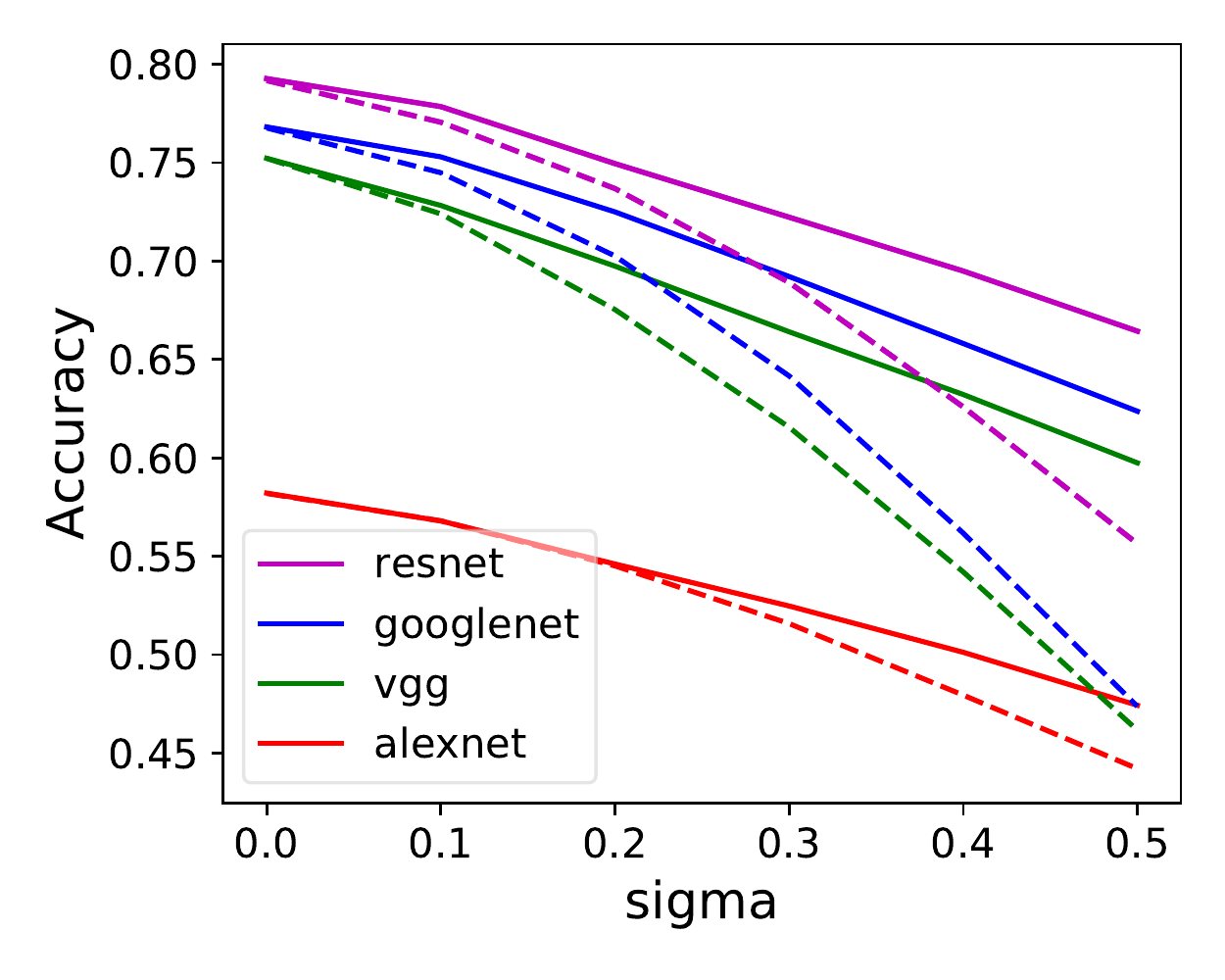}
        \caption{Setup 2 results}
        \label{fig:setup2}
    \end{subfigure}
    \hfill
    \begin{subfigure}[b]{0.32\textwidth}
        \centering
        \includegraphics[width=\textwidth]{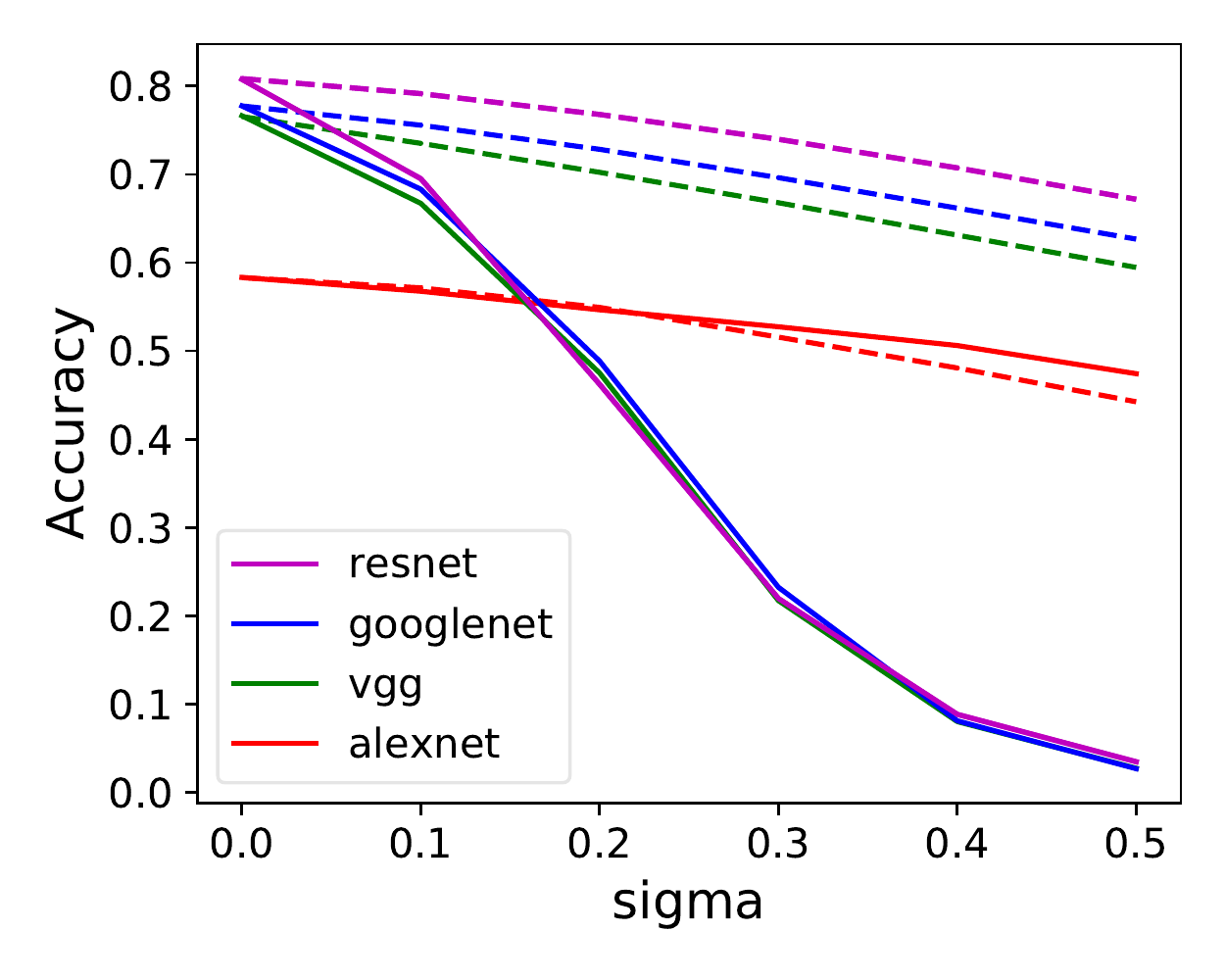}
        \caption{Setup 3 results}
        \label{fig:setup3}
    \end{subfigure}
    \caption{Effects of degradation on classification networks trained on three setups. The solid line means degraded images directly tested by classification networks, and the dashed line means degraded images are restored and then tested by classification networks.}
    \label{fig:effects}
\end{figure}

We implement additive white Gaussian noise (AWGN): we add white Gaussian noise to the original clean images, and the additive noise is independent among three color channels (RGB). 
For different degradation levels, we change the standard deviation (sigma) of the normal distribution of the noise. 
We also implement the other four types of degradation (Gaussian blur, motion blur, salt and pepper noise and rectangle crop) to evaluate their effects on CNN-based classification networks and report the results in Appendix~\ref{app:other-degradation}. 
For further test, we implement spatially varying degradation with 1D and 2D varying: the degradation level linearly changes with the number of rows or column randomly of original images for 1D varying; the degradation level linearly changes with respect to the Euclidean distance to a random point for 2D varying. Fig. \ref{fig:degradation} illustrates some examples of synthesized images of different degradation types and levels.
In the first setup, we directly use classification networks trained on clean images and test them on uniformly degraded images with different degradation levels and report the results in Fig. \ref{fig:setup1} in solid lines.

For AWGN, we find a corresponding state-of-art restoration network DnCNN\footnote{https://github.com/cszn/KAIR}~\cite{zhang2017beyond}  to evaluate the performance of low-level vision networks on high-level tasks. 
We train the DnCNN network from scratch and use the same training images and argumentation as that for fine-tuning classification networks except those images are degraded by our AWGN degradation model. 
We also use the same training procedure as for classification including batch-step linearly learning rate warmup and cosine learning rate decay. The difference is that we reduce the initial learning rate to 0.0001 with Adam optimizer, and increase the batch size to 128, with a patch size of 50-by-50, and a stride of 25. 
We also change the loss function $\ell_2$ in original paper~\cite{zhang2017beyond} to $\ell_1$ as it achieves better convergence performance~\cite{zhao2016loss}.
We restore the degraded images before testing them in classification networks. The results are shown in Fig. \ref{fig:setup1} in dashed lines.

\subsection{Setup 2: train on degraded images}
\label{sub:degraded}
In \cite{pei2019effects}, the authors noticed that training classification networks on degraded images can also improve their performance on degraded images, and the restoration network does not have improvements in this situation. Therefore, following their work, we also fine tune classification networks on mixture of degraded images with all discrete degradation levels given in the previous setup, including clean images (i.e. $\sigma\in\{0, 0.1, 0.2, 0.3, 0.4, 0.5\}$). We test those classification networks trained on degraded images with the same setting in Section~\ref{sub:clean} and report the results in Fig.~\ref{fig:setup2}.

\subsection{Setup 3: train on restored images}

The last setup is fine-tuning classification networks on mixture of degraded (as in Section~\ref{sub:degraded}) then restored images by pre-trained DnCNN (as in Section~\ref{sub:clean}), including clean images. 
We test those classification networks with the same settings as in the previous setups and report the results in Fig. \ref{fig:setup3}.

\section{Analysis}
\label{sec:analysis}

By summarizing and analyzing the results from the previous three setups in Fig.~\ref{fig:effects}, we can obtain some interesting conclusions.
It is obvious to find that with the increment of the degradation level, the classification accuracy decreases among all three setups, which is reasonable since more severe degradation leads to more information loss from original clean images. 
Therefore, it becomes harder for classification networks to give correct predictions. 

From setup 1 and Fig.~\ref{fig:setup1}, we can also find that restoration network can increase the classification accuracy of degraded images when  classification networks are trained on clean images.
This shows that the restoration network can help classification networks overcome the problem of degradation.

However, from setup 2 and Fig.~\ref{fig:setup2}, training classification network on mixed-level degraded images can greatly increases classification accuracy on degraded images, which indicates that the classification networks themselves can overcome the problem of degradation given degraded training data. In this situation, the restoration network hurts the final performance, which is also noticed in~\cite{pei2019effects}. This might be because the mechanism classification networks used to eliminate degradation is conflict with the restoration network.

For setup 3, the results are similar to those obtained in setup 1, and the performances are improved compared with setup 1. Therefore, using the restoration network as a pre-processing step at training phase has a positive effect on the final results. 

\begin{figure}
    \centering
    \begin{subfigure}[b]{\linewidth}
        \centering
        \includegraphics[width=\linewidth]{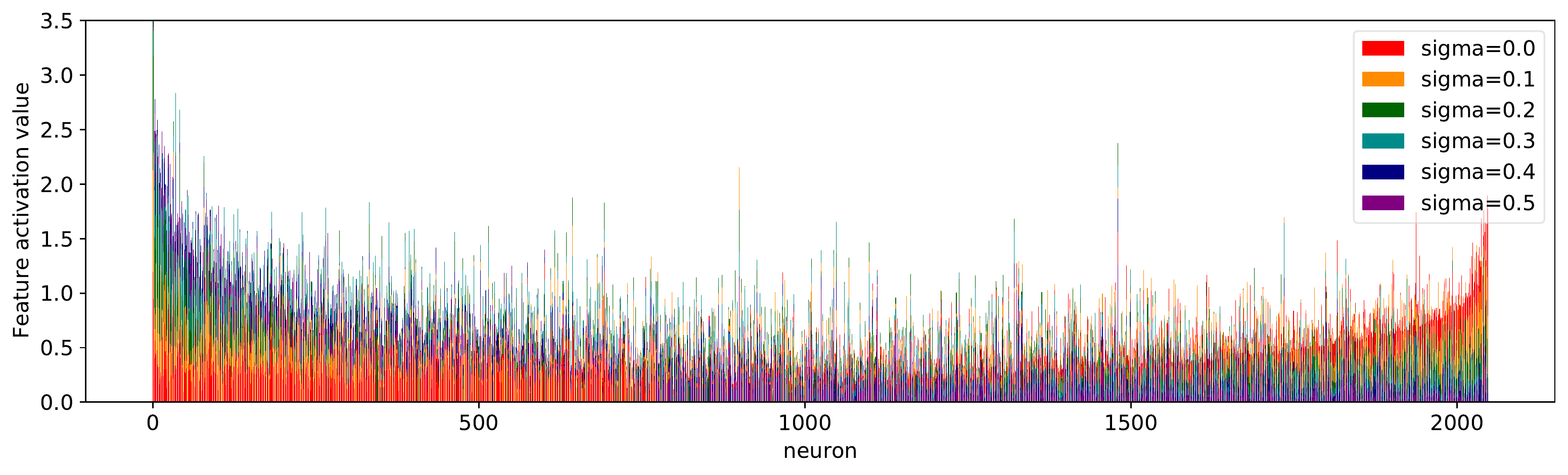}
        \caption{Before restoration}
    \end{subfigure}
    \hfill
    \begin{subfigure}[b]{\linewidth}
        \centering
        \includegraphics[width=\linewidth]{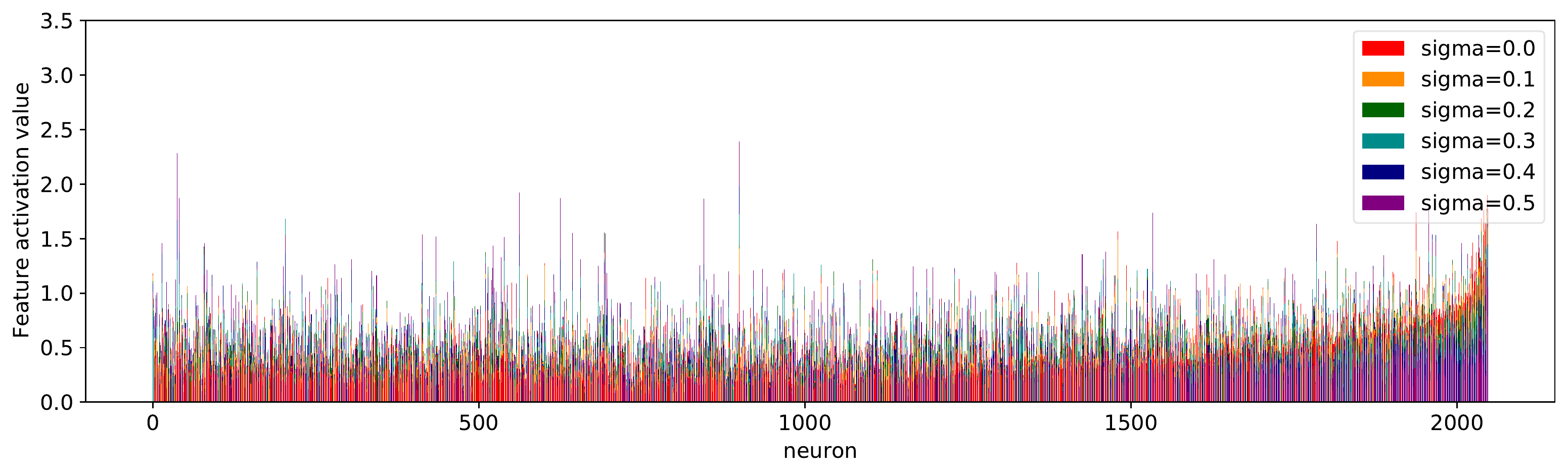}
        \caption{After restoration}
    \end{subfigure}
    \caption{Mean of activation values of each neuron (inputs of final FC layer) among images of a specific class (’basketball-hoop’).  We order neurons according to the difference between the value from clean and the most severely degraded images.}
    \label{fig:activation}
\end{figure}

For an in-depth analysis of the effects of degradation on classification network predication, we visualize the feature activation of different degradation levels on validation set. Fig.~\ref{fig:activation} illustrates the inputs of the final fully-connected layer in the ResNet-50 classification network. We choose a specific label (\textit{basketball-hoop}) and collect feature activation at different degradation levels. The comparison between those two figures further proves that the restoration network can help classification networks eliminate the influence from image degradation.

\chapter{Proposed method}
    In this chapter, we first introduce the fidelity map in Section \ref{sec:fidelitymap}, as well as some background of image classification networks and the overall pipeline of our proposed method in Section \ref{sec:background}. In section \ref{sec:architecture}, we introduce the model to use fidelity map in details. To stay focus on the proposed method and simplify our work, in this chapter and following two chapters, we will use ResNet-50~\cite{he2016deep} as an example to show the architecture and performance of the proposed method.

\section{Fidelity map}
\label{sec:fidelitymap}
The fidelity map is the pixel-wise $\ell_1$ or $\ell_2$ or cosine distance between restored images after a certain restoration network, like DnCNN, and the original clean images. 
We finally choose the $\ell_1$ distance for the proposed method, as it achieves the best performance.
The comparison of different fidelity map is given in Section \ref{sec:fidelity}.
Since our degraded images are synthesized from clean images by our degradation model, we can obtain the ground-truth fidelity in this situation (marked as Oracle below). In practice, we can also train a network as the fidelity map estimator.
In this report, we implement a fidelity map estimator using the same architecture as DnCNN.

\section{Overall pipeline}
\label{sec:background}
\begin{figure}
    \centering
    \includegraphics[width=.7\linewidth]{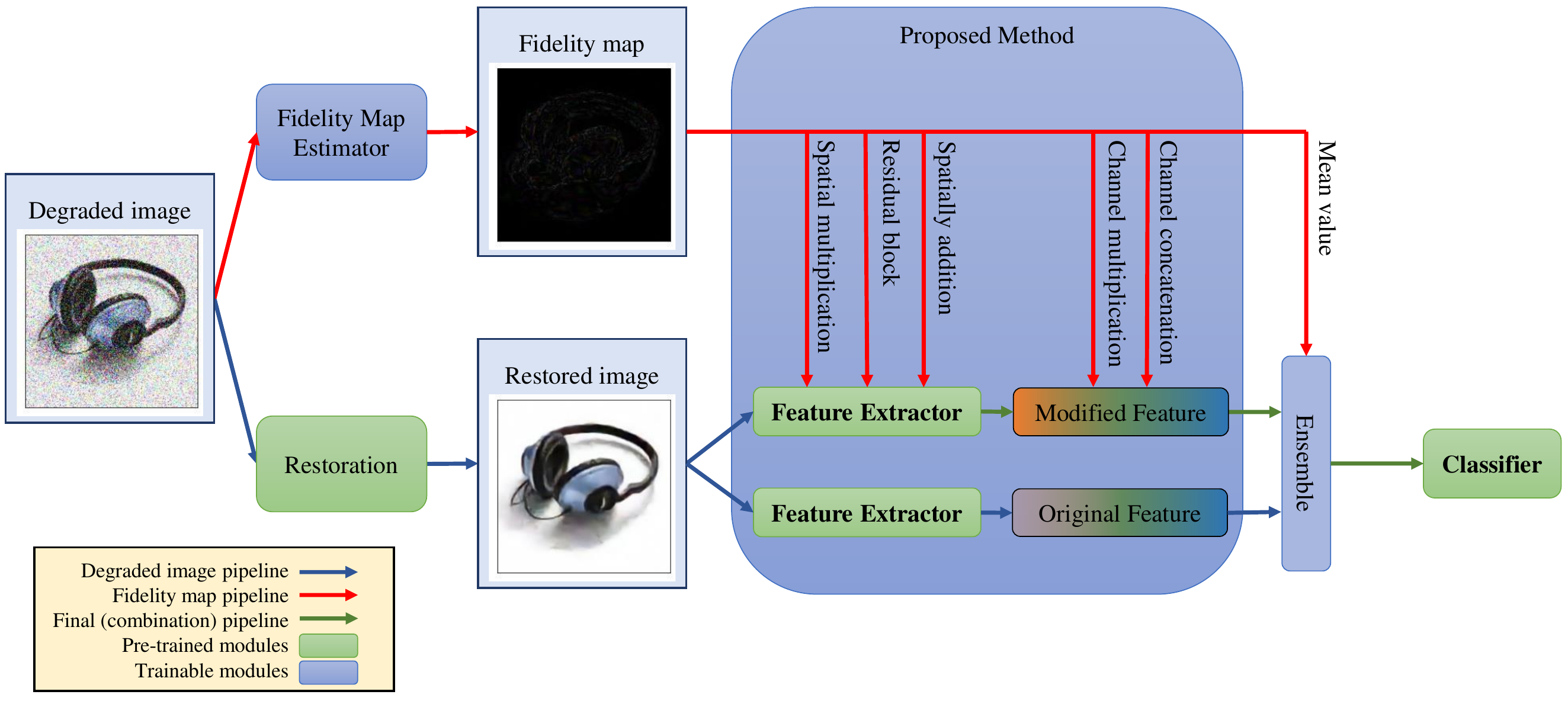}
    \caption{Overall pipeline of proposed method}
    \label{fig:pipeline}
\end{figure}

Existing CNN-based image classification networks typically can be split into two modules: a \textbf{feature extractor} to obtain the high-level feature from an image and a \textbf{classifier} which mainly consists of fully-connected layers for classification. 
The purpose of our method is to help the pre-trained feature extractor overcomes degradation by modifying the outputs or inputs of some layers in it. Therefore, we use classification networks pre-trained on clean images (the same as setup 1 in Section~\ref{sub:clean}).
The overall pipeline of our method is summarized in Fig. \ref{fig:pipeline}.

\section{Detailed architecture}
\label{sec:architecture}
\begin{figure}
    \centering
    \includegraphics[width=\linewidth]{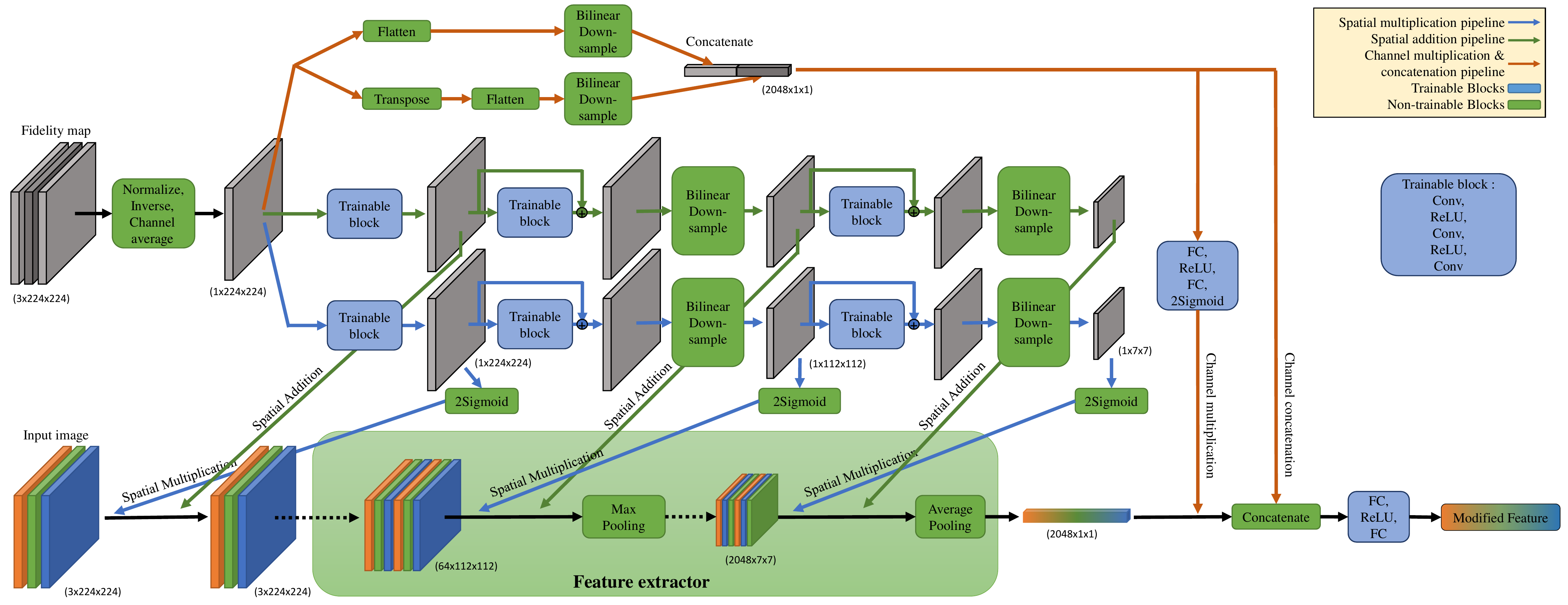}
    \caption{Detailed architecture (except ensemble block) of the proposed method based on ResNet-50~\cite{he2016deep}.}
    \label{fig:architecture}
\end{figure}

In this section, we introduce the proposed method in details. The proposed model mainly contains six modules: spatial multiplication, spatial addition, channel multiplication, channel concatenation, ensemble module and residual mechanism. The detailed architecture of the proposed method (except ensemble block) is shown in Fig. \ref{fig:architecture}.

\subsection{Spatial multiplication}
To overcome the degradation problem, the feature extractor should pay less attention to  more noisy pixels. 
As defined in the previous section, if we normalize (see Appendix~\ref{app:fidelity-normalizaion} for more details) the fidelity map and take the inverse of it, the resulted fidelity map should indicate the confidence level per pixel. 
Therefore, the most intuitive method is to element-wisely multiply the resulted fidelity map with the restored images and outputs of each layer of the feature extractor. 

Since the CNN structure in feature extractor keeps the spatial information of original images, we can directly apply point-wise multiplication for the corresponding location and broadcast the manipulation among all channels of the output of CNN. 
We add a trainable layer (Conv+ReLU+Conv+ReLU+Conv) in between two sequential spatial multiplication operations. 
In this situation, we use the fidelity map as an attention mechanism on the feature space, so we add a \textit{Sigmoid} operation before multiplication. 
The results of \textit{Sigmoid} only take value in $(0,1)$, which means we can only eliminate the value in feature space. However, the noise can also reduce the activation values, in this situation, the fidelity map makes the performance worse. To solve the problem, we use 2 times \textit{Sigmoid} results to make the activation values can be either increased or decreased.

Applying this operation before every layer of the feature extractor is not necessary and makes the whole model very complex. 
We only apply it on the inputs of the feature extractor and before every \textit{Pooling} layer inside the feature extractor.

\subsection{Residual mechanism} 
The residual network \cite{he2016deep} explicitly learns a residual mapping for a trainable layer. It is originally proposed to solve the performance saturation on very deep CNN models. Although the proposed model is not very deep, we still utilize the residual learning mechanism, as it shows significant improvements in the denoising networks \cite{zhang2017beyond} and other image restoration networks.

\subsection{Spatial addition}
According to the definition of fidelity map, when the fidelity map is $\ell_1$ distance, we can also add the fidelity map feature directly on the image feature. For this spatial addition module, we use the same architecture as spatial multiplication. 
The only difference is that we add the fidelity feature point-wisely among all channels instead of multiply, and we also remove the \textit{Sigmoid} operation since this is no longer an attention mechanism.

\subsection{Channel multiplication}
The above-mentioned modules are all implemented spatially, which means for all channels, they execute the same operation. 
However, the feature extractor also stores different image feature information in different channels. 
For example, the output of the feature extractor of ResNet-50~\cite{he2016deep} has 2048 channels. 
The image feature channel is increased from 3(RGB) to 2048 by a stack of convolutional layers. 
We can also train a similar feature extractor for fidelity map to get a feature with 2048 channels, but in that case, the complexity of proposed method will be the same as retraining a classification network on degraded images, which is not what we expected.

For channel operation, we flatten the fidelity map and downsample it to the proper length. Because the flatten operation will make use lose 2-D spatial information, we apply the same operation on the transposed fidelity and concatenate the two downsampled fidelity maps together as our fidelity feature. 
Therefore, the length of the downsampled fidelity map is the same as the number of channels of image features obtained by the original feature extractor.

Like spatial multiplication, we also pass the fidelity map feature through several trainable layers (FC+ReLU+FC) and apply point-wise multiplication with the image feature. We also introduce the \textit{Sigmoid} operator as in spatial multiplication.

\subsection{Channel concatenation}
We also concatenate the fidelity feature and image feature together and then pass the concatenated feature (size of $2\times2048$ for ResNet-50) into several trainable layers (FC+ReLU+FC). 
The final feature after the proposed method is of the same size as the original feature. 

\subsection{Ensemble module}
Finally, we ensemble the modified feature from the proposed method and the original image feature obtained from feature extractor trained on clean images. 
We implement this ensemble mechanism for each element between the modified and original feature as shown in Fig.~\ref{fig:pipeline}.

\chapter{Experiments}
    In this chapter, we implement the proposed method on ResNet-50 and run a series of experiments to explore the performance of the proposed method. 
We use the same Caltech-256 dataset as used in Chapter~\ref{cha:problem} and the three setups as baselines for comparison to show the improvements of the proposed method.
In Section~\ref{sec:procedure}, we first introduce the training procedure of our model. 
Our experiment can mainly be divided into three categories, we introduce the details of those experiments in the following section. Finally, we make a comparison between those experiments and baseline methods.

\section{Training procedure}
\label{sec:procedure}

To train the proposed network, we need to obtain the pre-trained classification network on clean images and pre-trained DnCNN network. 
We follow exactly the same procedure as mentioned in Section~\ref{sub:clean} to get pre-trained networks.

For the proposed part, we follow very similar procedure, using the same batch-step linearly learning rate warmup \cite{goyal2017accurate} for the first 5 epochs and cosine learning rate decay \cite{he2019bag} as well as label smoothing \cite{Szegedy_2016_CVPR} with $\varepsilon=0.1$. We use Nesterov Accelerate Gradient (NAG) descent \cite{nesterov1983method} optimizer with an initial learning rate of 0.001 and a batch size of 64. The total number of training epochs is set to be 50, and we also choose the model with the highest accuracy on the validation set among all training epochs for further test and analysis.

To get the pre-trained fidelity map estimator, we also follow the same procedure as we train DnCNN for restoration except for the output of the network is $\ell_1$ distance between original clean images and the restored result from pre-trained DnCNN for restoration.

\subsection{Oracle fidelity map estimator}
Since the noisy images used here is generated by degradation model from clean images, we can obtain the ground-truth value of fidelity map. 
In this setting, we use the ground-truth fidelity map instead of pre-trained fidelity map estimator to train the proposed model on mixture of degraded images, and obtain the results with and without ensemble module. Results are shown in Table~\ref{tab:experiment} (marked as 'oracle').

\subsection{Pre-trained fidelity map estimator}
In this setting, we use the results from the pre-trained fidelity map estimator to train our proposed model. This is more like realistic scenarios where ground-truth clean images cannot be obtained. We also obtain the results with and without ensemble module and report them in Table~\ref{tab:experiment} (marked as 'pre-train').

\subsection{End-to-end-trained fidelity map estimator}
In the last setting, we train the fidelity map estimator and our model end-to-end, using the same procedure as training proposed model. 
However, from our observation, training the fidelity map estimator from scratch under this setting will get worse performance. 
This might because the fidelity map estimator is trained on small patches (50-by-50) instead of the whole image as classification networks.
To solve this problem, we reload the pre-trained fidelity map estimator before end-to-end training (i.e. end-to-end fine-tuning the fidelity map estimator), but our proposed model applied on feature extractor is still trained from scratch. 
We also obtain the results with and without ensemble module and report them in Table~\ref{tab:experiment} (marked as 'end-to-end').

\begin{table}[htbp]
  \footnotesize
  \centering
  \caption{Classification accuracy(\%) of the baseline methods (three setups in Section~\ref{sec:effect}) and the proposed methods using oracle, pre-trained and end-to-end-trained fidelity map estimator. Best results under different degradation levels and types are in bold.}
    \begin{tabular}{|c|c|c|c|c|c|c|c|c|c|c|}
    \hline
    \multicolumn{3}{|c|}{\multirow{2}[4]{*}{Methods}} & \multirow{2}[4]{*}{Clean} & \multicolumn{5}{c|}{Uniform degraded (sigma)}                  & \multicolumn{2}{c|}{Spatially } \\
    \cline{5-11}    
    
    \multicolumn{3}{|c|}{}  &   & 0.1   & 0.2   & 0.3   & 0.4   & 0.5   & 1D    & 2D \\
    \hline
    
    \multirow{2}[4]{*}{setup 1: train on clean} 
    & \multicolumn{2}{c|}{test w/o restoration} 
    & 83.20      & 62.80      & 30.24      & 10.44      & 3.73       & 1.65       & 28.60      & 32.72 \\
    \cline{2-11}              
    & \multicolumn{2}{c|}{test w/ restoration} 
    & \textbf{83.23}      & 77.38      & 68.20      & 56.05      & 42.40      & 30.32      & 65.65      & 67.78 \\
    \hline
    
    \multirow{2}[4]{*}{setup 2: train on degraded} 
    & \multicolumn{2}{c|}{test w/o restoration} 
    & 79.27      & 77.85      & 74.95      & 72.23      & 69.49      & 66.44      & 73.88      & 74.75 \\
    \cline{2-11}               
    & \multicolumn{2}{c|}{test w/ restoration} 
    & 79.17      & 77.05      & 73.68      & 68.91      & 62.59      & 55.64      & 71.83      & 72.95 \\
    \hline
    
    \multirow{2}[4]{*}{setup 3: train on restored} 
    & \multicolumn{2}{c|}{test w/o restoration} 
    & 80.76      & 69.51      & 46.22      & 21.96      & 8.84       & 3.47       & 43.23      & 47.44 \\
    \cline{2-11}               
    & \multicolumn{2}{c|}{test w/ restoration} 
    & 80.85      & 79.14      & 76.78      & 73.97      & 70.73      & 67.18      & 75.78      & 76.36 \\
    \hline
    
    \multirow{6}[12]{*}{Proposed} 
    & \multirow{2}[4]{*}{oracle} 
    & w/o ensemble & 82.29      & 80.24      & 78.06      & 76.24      & 74.36      & \textbf{72.38}      & 77.65      & 78.03 \\\cline{3-11}
    & & w/ ensemble & 82.37      & \textbf{80.43}      & \textbf{78.17}      & \textbf{76.30}      & \textbf{74.43}      & 72.35      & \textbf{77.89}      & \textbf{78.25} \\\cline{2-11}
    & \multirow{2}[4]{*}{pre-trained} 
    & w/o ensemble & 82.77      & 79.01      & 74.78      & 69.86      & 64.66      & 58.73      & 73.59      & 74.46 \\\cline{3-11}               
    & & w/ ensemble & 83.13      & 79.34      & 75.21      & 70.16      & 64.62      & 58.40      & 73.97      & 74.81 \\\cline{2-11}               
    & \multirow{2}[4]{*}{end-to-end} 
    & w/o ensemble & 83.04      & 79.54      & 75.45      & 70.40      & 64.98      & 58.99      & 73.88      & 75.06 \\\cline{3-11}
    &  & w/ ensemble & 83.22      & 79.56      & 75.57      & 70.49      & 64.92      & 58.73      & 74.10      & 75.19 \\
    \hline
    \end{tabular}%
  \label{tab:experiment}%
\end{table}%

\section{Comparison between proposed method and baseline methods}

By comparing and analyzing the results in Table~\ref{tab:experiment}, we find that the proposed oracle model achieves the best results under degradation. However, since we use ground-truth clean images in this setting, this is an unfair comparison. These results show that given a more powerful fidelity map estimator, our model has the potential to get better performance in a realistic situation.

Our proposed model uses the same pre-trained classification networks as baseline setup 1: train on clean. By comparing, we find that our proposed method under all three settings outperforms baseline setup 1. As degradation becomes more severe, our improvements become more significant. This shows that the proposed method indeed can help pre-trained classification networks perform better on degraded images. 

When comparing the proposed method with other baseline methods, we find that the proposed method gets better performance on more slightly degraded images. As degradation becomes more severe, training classification networks directly on degraded images gets better results. However, this method retrains the whole classification network leading the worst performance on clean images. 

We also provide the results of proposed results with and without ensemble.
The ensemble module improves the accuracy on clean and slightly degraded images at the price of a small decrease of classification accuracy on severely degraded images. This is because the ensemble module introduces original images feature which performs better on clean images.

\chapter{In-depth analysis \& discussion}
    In this chapter, we run a series of experiments and conduct an in-depth analysis of the results on the oracle model without ensemble to show the improvement caused by each module in the proposed method.

\section{In-depth analysis}
\subsection{Different fidelity map}
\label{sec:fidelity}
As mentioned in Section~\ref{sec:fidelitymap}, we can also define the fidelity map as pixel-wise $\ell_2$ or cosine distance between restored images and the original clean images. We implement these two types of fidelity map and keep other parts the same as proposed oracle method without ensemble. The results are recorded in Table. \ref{tab:ablation} and show that $\ell_1$ distance can achieve the best performance among all three types of fidelity map.
\begin{table}[htbp]
  \small
  \centering
  \caption{In-depth analysis and ablation study results (classification accuracy in \%). Test are done on oracle model without ensemble. Best results under different degradation levels and types are in bold.}
    \begin{tabular}{|c|c|c|c|c|c|c|c|c|}
    \hline
    \multicolumn{2}{|c|}{\multirow{2}[4]{*}{Methods}} & \multirow{2}[4]{*}{Clean} & \multicolumn{5}{c|}{Uniform degraded (sigma)}                  & Spatially  \\
\cline{4-9}    \multicolumn{2}{|c|}{}  &            & 0.1        & 0.2        & 0.3        & 0.4        & 0.5        & 1D \\
    \hline
    \multicolumn{2}{|c|}{porposed method: oracle w/o ensemble} & 82.29      &80.24      & 78.06      & \textbf{76.24}      & \textbf{74.36}      & \textbf{72.38}      & 77.65 \\
    \hline
    \multicolumn{2}{|c|}{setup 1: train on clean test w/ restoration} & \textbf{83.23}      & 77.38      & 68.20      & 56.05      & 42.40      & 30.32      & 65.65 \\
    \hline
    \multirow{2}[4]{*}{Fidelity map} & $\ell_2$ distance & 82.57      & 80.15      & \textbf{78.16}      & 76.21      & 74.30      & 72.00      & 77.80 \\
\cline{2-9}               & cosine     & 82.56      & 79.53      & 75.11      & 69.54      & 62.96      & 55.70      & 73.72 \\
    \hline
    \multirow{2}[4]{*}{Downsampling} & bicubic    & 82.39      & \textbf{80.38}      & 78.15     & 76.19      & 74.26      & 72.07      & \textbf{77.76} \\
\cline{2-9}               & nearest    & 82.52      & \textbf{80.38}      & 78.02      & 76.18      & 74.22      & 71.97      & 77.71 \\
    \hline
    \multirow{5}[10]{*}{Ablation study} & w/o spatial multiplication & 82.56      & 80.30      & 77.70      & 75.38      & 72.95      & 70.29      & 77.16 \\
\cline{2-9}               & w/o residual mechanism & 81.80      & 79.80      & 77.55      & 75.53      & 73.35      & 70.99      & 76.96 \\
\cline{2-9}               & w/o spatial  addition & 82.61      & 79.76      & 77.00      & 74.11      & 71.44      & 67.43      & 76.19 \\
\cline{2-9}               & w/o channel multiplication & 82.42      & 80.32      & 78.10      & 75.97      & 74.15      & 71.98      & 77.47 \\
\cline{2-9}               & w/o channel concatenation & 82.13      & 79.61      & 76.57      & 73.99      & 71.50      & 68.85      & 76.19 \\
    \hline
    \end{tabular}%
  \label{tab:ablation}%
\end{table}%

\subsection{Different downsampling methods}
Since we use downsampling in the proposed method and there are many downsampling methods, here we choose other two popular used methods (bi-cubic and nearest) in computer vision tasks for comparison. The results are recorded in Table. \ref{tab:ablation}. We find that bi-linear downsampling performs better under more severe degradation, and nearest and bi-cubic perform better on clean images and slightly degraded images respectively. Because our purpose is to improve accuracy under degradation and we can make up the performance on clean images by using the ensemble module, we choose bi-linear downsampling for our method.  

\section{Ablation study}
To explore the contribution of each module introduced in Section~\ref{sec:architecture}. We designed experiments for ablation study. For each experiment, we only remove one module from our full model and record the result in Table. \ref{tab:ablation}. The results show that removing any module will cause worse accuracy which indicates all of modules contribute to the final best performance.

\chapter{Generalization ability studies}
\begin{figure}
    \centering
    \begin{subfigure}[b]{.8\linewidth}
        \centering
        \includegraphics[width=\linewidth]{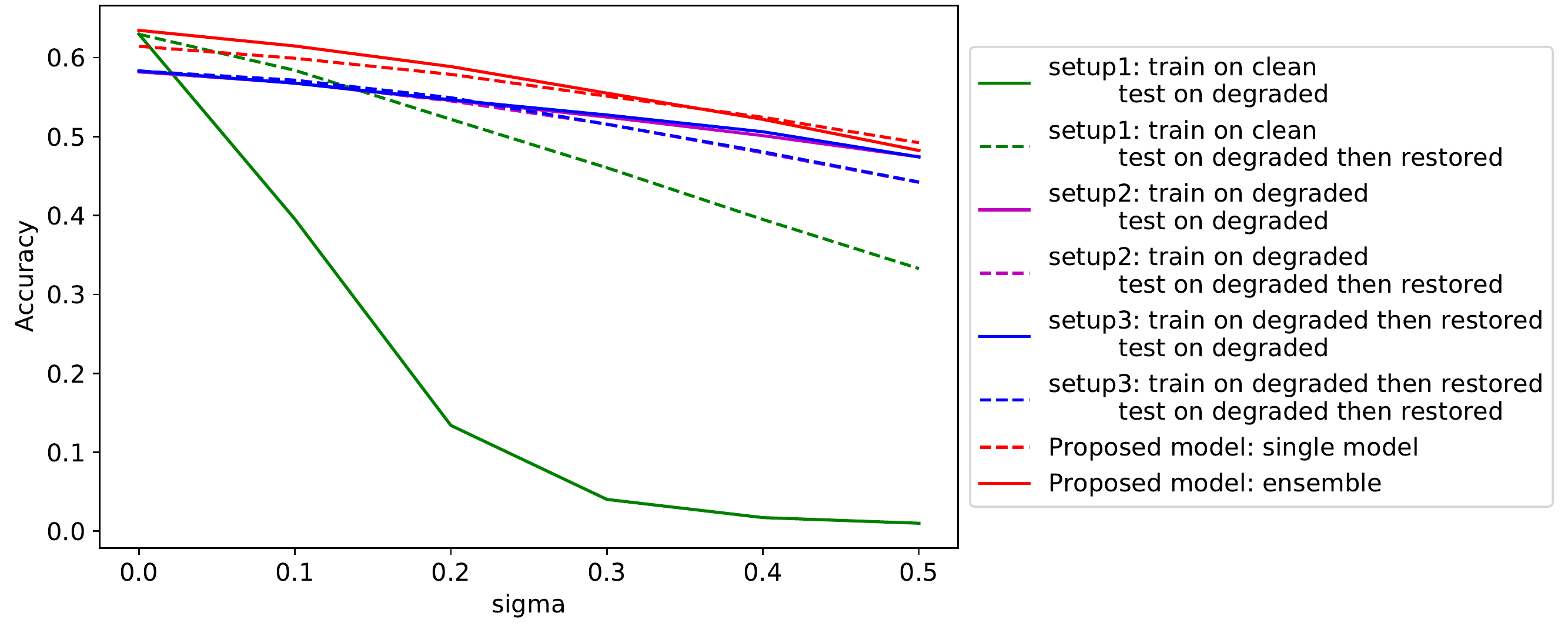}
        \caption{AlexNet}
        \label{fig:generalibity-alex}
    \end{subfigure}
    
    \vspace{20pt}
    \begin{subfigure}[b]{.8\linewidth}
        \centering
        \includegraphics[width=\linewidth]{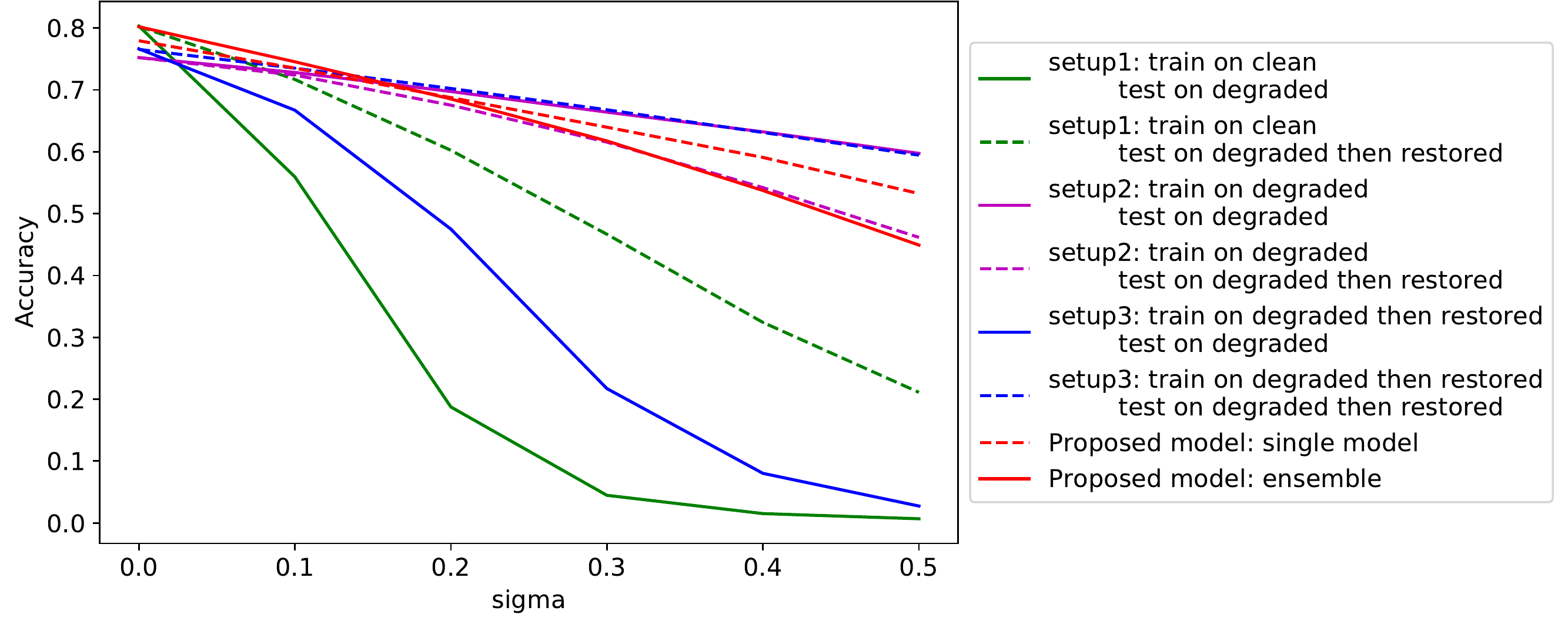}
        \caption{VGG}
        \label{fig:generalibity-vgg}
    \end{subfigure}
    
    \vspace{20pt}
    \begin{subfigure}[b]{.8\linewidth}
        \centering
        \includegraphics[width=\linewidth]{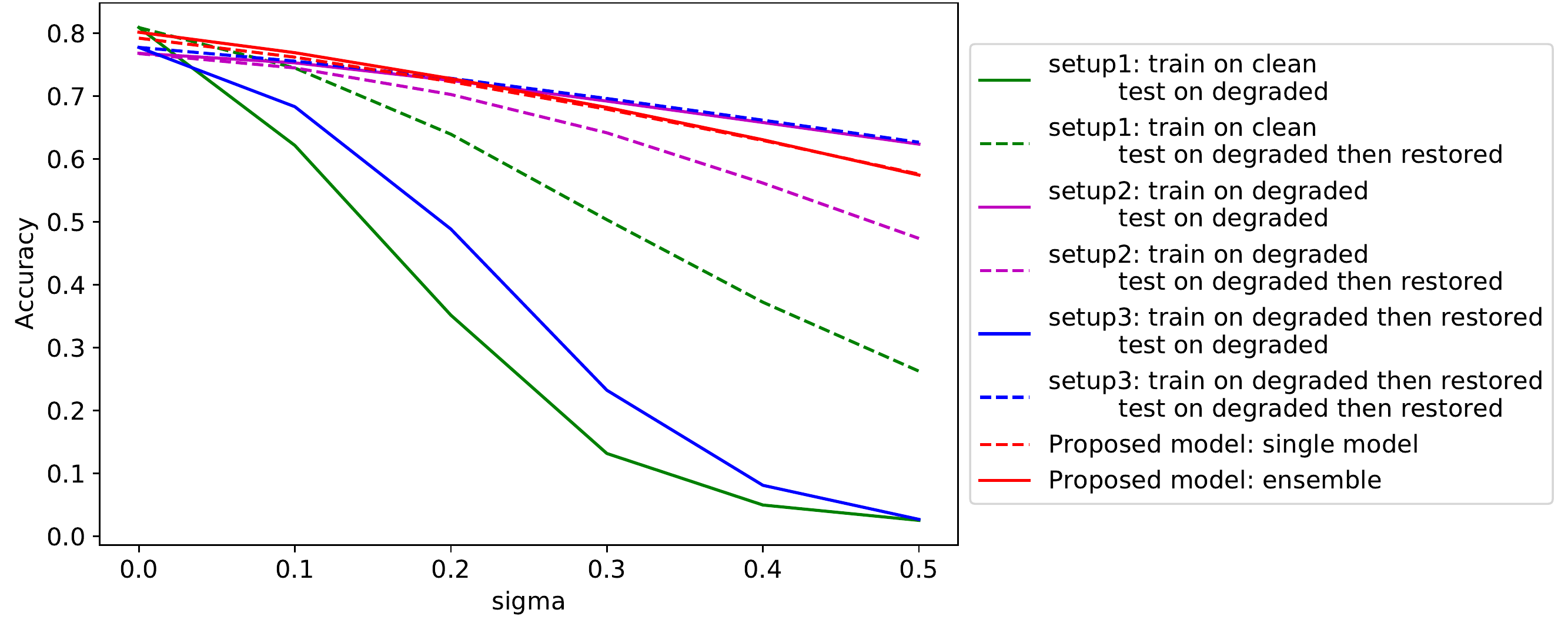}
        \caption{GoogLeNet}
        \label{fig:generalibity-goo}
    \end{subfigure}
    \caption{Performance of proposed methods and baseline methods on other classification networks.}
    \label{fig:generalibity}
\end{figure}

    In this chapter, we implement the proposed method on the other three classification networks to show the generalization ability of the proposed method . We run the proposed method under end-to-end setting, and the results are illustrated in Fig.~\ref{fig:generalibity}.

From Fig.\ref{fig:generalibity-vgg} and Fig.\ref{fig:generalibity-goo}, we find the performances of the proposed method on VGG and GoogLeNet are similar to that on ResNet-50, which demonstrates the proposed method has strong generalization ability to be used by other classification networks. 
We are surprised to find that the proposed method achieves the best result on AlexNet. We think that is because AlexNet uses more \textit{MaxPooling} layers which leads to a deeper architecture for the proposed model. 

\chapter{Conclusion}
    In this report, we firstly explore the effects of degradation types and levels on classification networks. 
We conclude that the increments of the degradation level leads to the decrements of classification accuracy, and training classification network on mixed-level degraded images can greatly increase classification accuracy on degraded images.
Although restoration network can improve the classification accuracy of degraded images when classification networks are trained on clean images, it hurts the final performance when classification networks are trained on mixed-level degraded images.

We also propose a method making use of fidelity map which is the pixel-wise $\ell_1$ distance between restored images and clean images. 
We shows that the proposed method can greatly improve the performance of pre-trained classification networks on degraded images, especially under lower degradation level. 
We empirically demonstrate that all designed modules in the proposed method has positive effect on our improvements.

Finally, extensive empirical studies show that the proposed method is a model-agnostic approach that benefits other classification networks. However, we only test the generalization ability on other classification networks instead of other restoration networks or degradation types, which remains our future research.

\clearpage
\pagestyle{numberonly}
\renewcommand*{\bibname}{References}  
\printbibliography

\newpage
\appendix

\chapter{Effects of other degradation on classification networks}
\label{app:other-degradation}
\section{Degradation model}
\begin{figure}
    \centering
    \includegraphics[width=\linewidth]{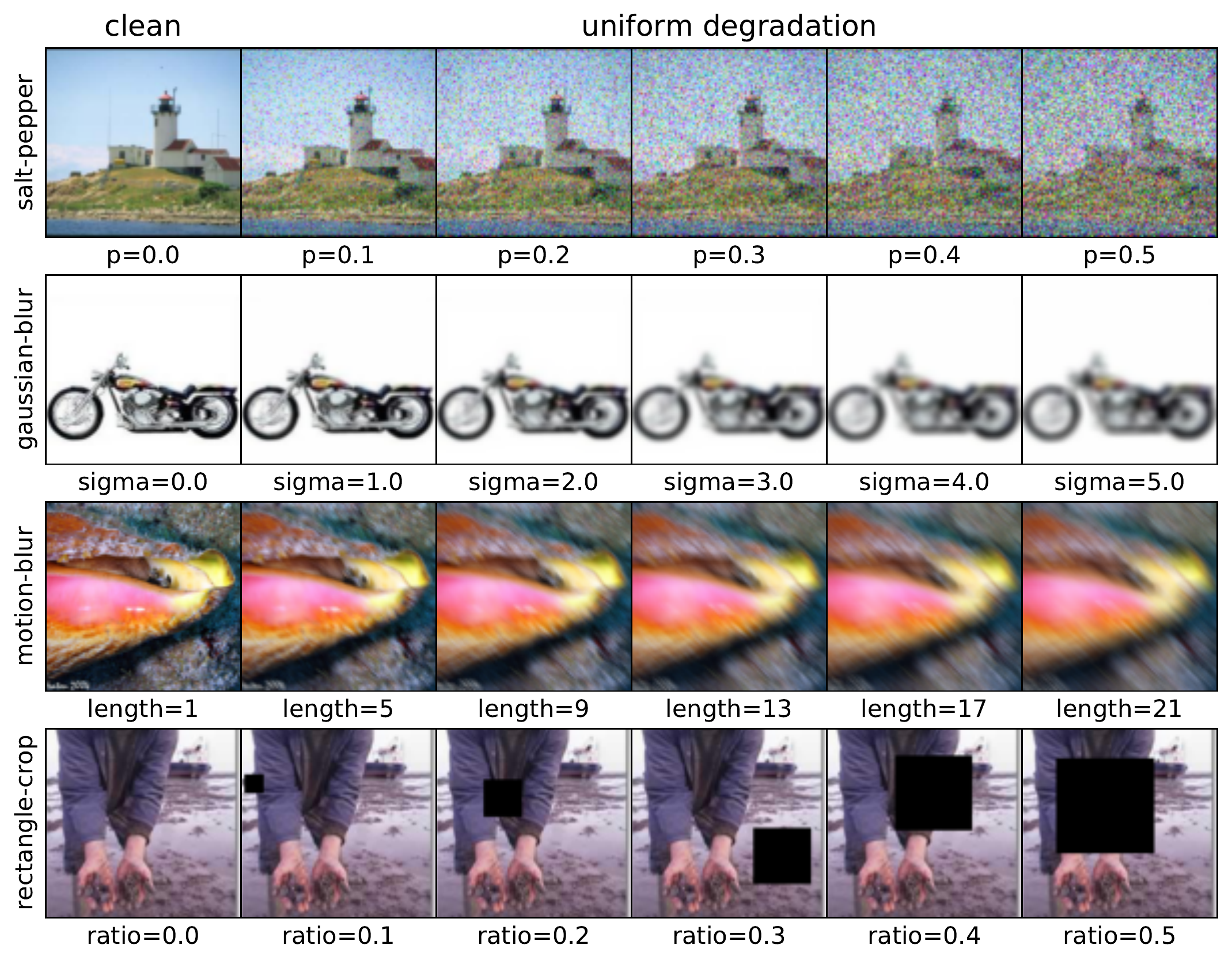}
    \caption{Examples of the synthesized degraded images with other degradation types.}
    \label{fig:degradation-app}
\end{figure}

To explore the effects of degradation types and levels on classification networks, we also implement four other types of degradation model: Gaussian blur, motion blur, salt and pepper noise, rectangle crop.  
\textbf{Gaussian blur}: we convolve the clean images with Gaussian kernel of size 13 following \cite{flusser2015recognition, pei2019effects}. The degradation level is changed by using a different value of standard deviation (sigma) of the Gaussian kernel.
\textbf{Motion blur}: the implementation of motion blur is similar to that of Gaussian blur, the only difference is using motion kernel instead of the Gaussian kernel. We implement motion kernel following \cite{pei2019effects, sun2015learning}, and the degradation level is controlled by varying the length of the blur kernel, and we consider only $45^o$ motion orientation.
\textbf{Salt and pepper noise}: we add this noise by replacing the pixels in original images with white or black pixels (channel-independent). The degradation level is changed by varying the probability of replacement. 
\textbf{Rectangular crop}: we crop a square from the original images. The degradation level is controlled by the ratio of the side length of the cropped square to the minimum side length of the original image.
Fig. \ref{fig:degradation-app} illustrates some examples of synthesized images of different degradation types and levels.

\section{Results of effects on classification networks}

\begin{figure}
    \centering
    \includegraphics[width=\linewidth]{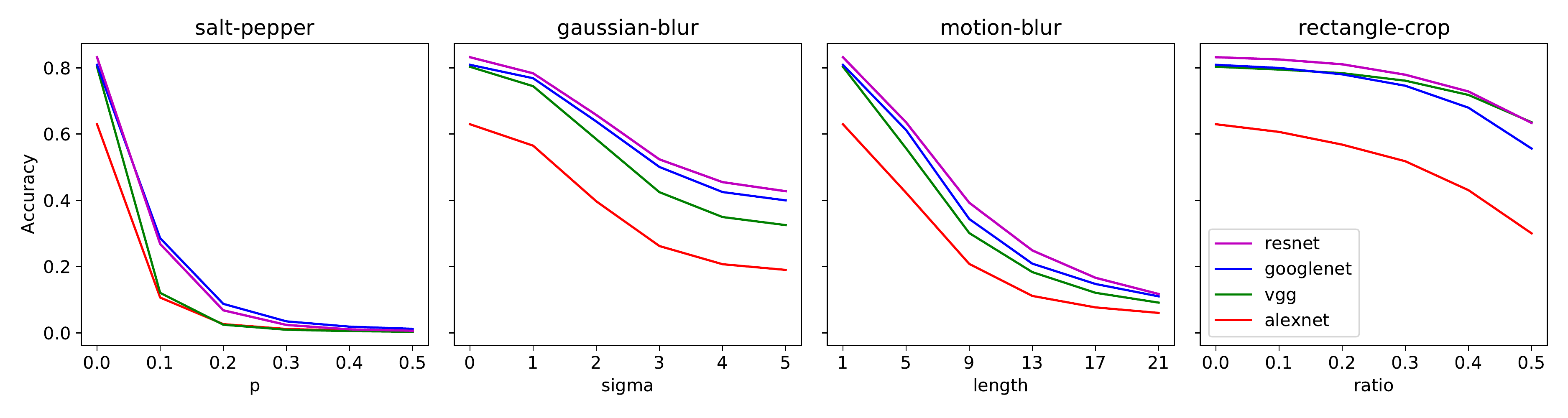}
    \caption{Performance of classification networks on other degradation types.}
    \label{fig:effects-clean-app}
\end{figure}

Following the same experiment setup 1 mentioned in Section~\ref{sub:clean}. We repeated the same experiment on our four classification networks on different degradation type except for AWGN and show the results in Fig~\ref{fig:effects-clean-app}.

\chapter{Fidelity map normalization}
\label{app:fidelity-normalizaion}
For $\ell_1$ and $\ell_2$ fidelity map, we apply normalization on the fidelity map before all of other manipulations. Since the noise follows normal distribution, and our train images combines images of different degradation levels $\sigma\in\{0, 0.1, 0.2, 0.3, 0.4, 0.5\}$, we assume the overall noise follows normal distribution with zero expectation defined by:
$$\sigma^2=(0.1^2+0.2^2+0.3^2+0.4^2+0.5^2)/6^2
$$
After passing through the restoration network, we assume $\sigma^2$ is halved, therefore:
$$\sigma^2=(0.1^2+0.2^2+0.3^2+0.4^2+0.5^2)/(6^2\times2)
$$
For $\ell_1$ fidelity map, the absolute of normal distribution follows a half-normal distribution with mean of $\frac{\sigma\sqrt{2}}{\sqrt{\pi}}$ and variance of $\sigma^2(1-\frac{2}{\pi})$. 
For $\ell_2$ fidelity map, the square of normal distribution follows Gamma distribution with mean of $\sigma^2$ and variance of $\sigma^2\sqrt{2}$.

\chapter{Experiment record}
\section{Fine tuning record of classification networks on clean images}

\begin{figure}[h]
    \centering
    \includegraphics[width=\linewidth]{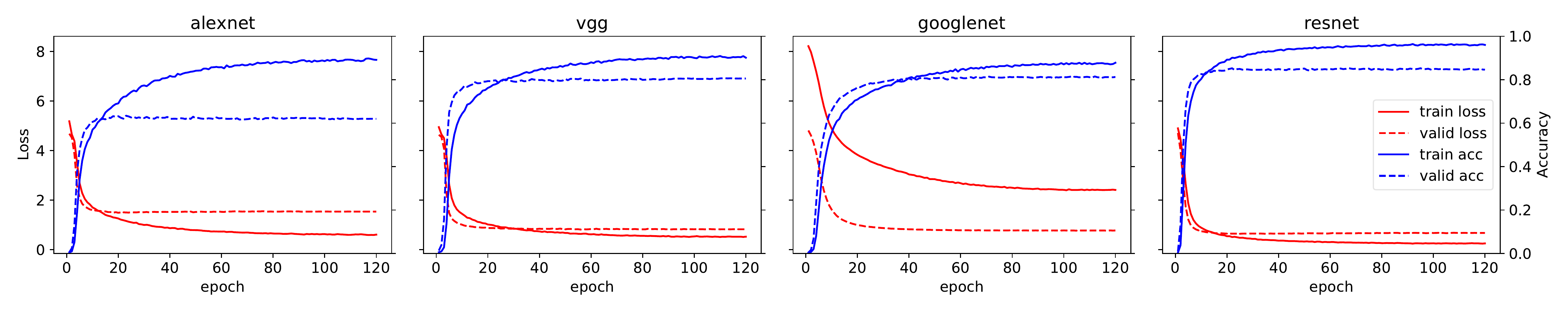}
    \caption{Loss and accuracy on train and validation set vs. number of fine tuning epochs.}
    \label{fig:finetune-clean}
\end{figure}

\typeout{get arXiv to do 4 passes: Label(s) may have changed. Rerun}
\end{document}